# EcomEval: Towards Reliable Evaluation of Large Language Models for Multilingual and Multimodal E-Commerce Applications

Shuyi Xie[*,1], Ziqin Liew[*,1], Hailing Zhang[*,1], Haibo Zhang[1], Ling Hu[1], Zhiqiang Zhou[1], Shuman Liu[1] and Anxiang Zeng[1]

[*]Equal contributions, [1]Shopee

Large Language Models (LLMs) excel on general-purpose NLP benchmarks, yet their capabilities in specialized domains remain underexplored. In e-commerce, existing evaluations—such as EcomInstruct, ChineseEcomQA, eCeLLM, and Shopping MMLU—suffer from limited task diversity (e.g., lacking product guidance and after-sales issues), limited task modalities (e.g., absence of multimodal data), synthetic or curated data, and a narrow focus on English and Chinese, leaving practitioners without reliable tools to assess models on complex, real-world shopping scenarios. We introduce EcomEval, a comprehensive multilingual and multimodal benchmark for evaluating LLMs in e-commerce. EcomEval covers six categories and 37 tasks (including 8 multimodal tasks), sourced primarily from authentic customer queries and transaction logs, reflecting the noisy and heterogeneous nature of real business interactions. To ensure both quality and scalability of reference answers, we adopt a semi-automatic pipeline in which large models draft candidate responses subsequently reviewed and modified by over 50 expert annotators with strong e-commerce and multilingual expertise. We define difficulty levels for each question and task category by averaging evaluation scores across models with different sizes and capabilities, enabling challenge-oriented and fine-grained assessment. EcomEval also spans seven languages—including five low-resource Southeast Asian languages—offering a multilingual perspective absent from prior work. We evaluate 19 open and proprietary LLMs on EcomEval, revealing substantial performance disparities and highlighting scenarios where these general-purpose models perform poorly in the e-commerce domain. By combining diversity, authenticity, quality, difficulty awareness, multilinguality and multimodality, EcomEval establishes a rigorous and representative testbed for advancing research and deployment of LLMs in e-commerce. Upon acceptance, we will release the full dataset [a] to support reproducible research.

[a]https://github.com/ShopeeLLM/EcomEval

## 1. Introduction

Large Language Models (LLMs) have achieved remarkable performance across general NLP benchmarks, but evaluating them in specialized domains like e-commerce remains challenging. Numerous benchmarks target broad capabilities – from knowledge-intensive QA to multi-task reasoning – providing ample evaluation data in generic settings. However, in the e-commerce domain, such dedicated benchmarks are sparse (Chen et al., 2025; Jin et al., 2024). This scarcity leaves a gap in assessing whether advanced LLMs possess the nuanced skills required for e-commerce applications. Practitioners currently lack reliable yardsticks to measure model competence on complex shopping-related queries, making it difficult to gauge how well these models handle real-world e-commerce tasks.

Recent efforts have started to explore e-commerce-specific LLM benchmarks, but each comes with limitations. Notable examples include EcomInstruct (Li et al., 2023), ChineseEcomQA (Chen et al., 2025), eCeLLM (Peng et al., 2024), and Shopping MMLU (Jin et al., 2024). EcomInstruct and eCeLLM introduced large instruction-tuned datasets for e-commerce, covering tasks like product description generation, attribute extraction, and recommendation. However, these rely heavily on synthetic or curated instruction data, which may not capture the full complexity of real user interactions. ChineseEcomQA focuses on fundamental concept question-answering in the e-commerce domain, but it is restricted to QA pairs (primarily in Chinese) and does not encompass the rich variety of e-commerce tasks. Shopping MMLU provides a broader multi-task evaluation with 57 tasks derived from Amazon data, covering areas such as concept understanding and knowledge reasoning. Yet, its format (largely multiple-choice questions) simplifies many challenges, and like the others, it overrepresents English and Chinese content. Common shortcomings across these benchmarks include limited task diversity and a reliance on instruction-tuned or synthetic data. Important real-world scenarios – for example, interactive product guidance dialogues or after-sales customer service queries – remain largely absent from existing evaluations. These issues underscore the need for a more comprehensive benchmark, one that spans diverse tasks, languages, modalities and realistic e-commerce complexities. Our work aims to fill this critical gap.

In this paper, we present EcomEval, a new benchmark that delivers a comprehensive and realistic evaluation suite for LLMs in the e-commerce domain. First, it achieves broad task diversity and multimodality, encompassing six primary categories and 37 distinct tasks that reflect genuine business needs, ranging from customer product question answering and attribute reasoning to intent understanding from queries or reviews and multi-turn shopping-assistant dialogues. Second, the benchmark is grounded in authentic data: most items are derived from real user queries and transaction logs, rather than purely synthetic or instruction-tuned sources, thereby capturing the complexity and noise inherent in real customer–merchant interactions. Third, it ensures high quality and scalability through a semi-automatic construction pipeline in which large language models generate initial drafts that are subsequently refined by over 50 expert annotators with strong e-commerce and multilingual expertise, balancing fast production with rigorous quality control. Fourth, the benchmark is explicitly challenge-oriented, incorporating calibrated difficulty levels at both task and item granularity, enabling fine-grained discrimination of model capabilities and surfacing performance gaps across systems. Finally, it supports multilingual evaluation, covering seven languages—including English, Chinese, and five additional Southeast Asian languages—thus moving beyond the English/Chinese focus of prior work and reflecting the truly global nature of e-commerce. By integrating diverse task coverage, authentic data, quality-assured scalability, difficulty-aware design, and multilingual breadth, our benchmark offers a rigorous and representative testbed for advancing LLM research and deployment in e-commerce.

We validate the effectiveness of our benchmark through extensive experiments. A total of 19 state-of-the-art LLMs – both open-source models and commercial systems – were evaluated on the full range of tasks. The results demonstrate that our benchmark robustly differentiates model capabilities, revealing significant performance disparities that were obscured by earlier, narrower evaluations. No single model dominated across all tasks, and even generally strong LLMs struggled with certain e-commerce challenges. For instance, models with impressive general NLP performance often faltered on tasks requiring nuanced understanding of domain-specific terminology or multi-step reasoning based on product knowledge. In particular, we observed that some top-performing general models (e.g., GPT-4o-tier systems) underperform on complex tasks like cross-lingual product QA and conversational recommendation, highlighting clear performance gaps. These findings are in line with recent observations that domain-tuned models can surpass even the best general LLMs on specialized e-commerce tasks. By exposing where current models excel and where they fail, the benchmark provides actionable insights into the strengths and weaknesses of today's LLMs in e-commerce. This differentiation underlines the benchmark's value in driving progress: it pinpoints which e-commerce scenarios remain most challenging for LLMs, guiding researchers and practitioners toward targeted improvements.

Main Contributions. We advance LLM evaluation for e-commerce through three key contributions:

- Comprehensive, Authentic, Multilingual and Multimodal E-commerce Classification System: We introduce a comprehensive classification system for e-commerce tasks, covering six categories and 37 diverse tasks (including 8 multimodal tasks), built from real e-commerce queries and transaction logs. It spans seven languages (English, Chinese, Indonesian, Vietnamese, Thai, Malay, Portuguese), addressing low-resource settings and reflecting the global breadth of online e-commerce.
- Open, High-Quality, Difficulty-Aware Dataset: We publicly release EcomEval ($\approx 3,100$ items) with reference answers and calibrated difficulty levels at both task and item granularity, enabling reproducible research and rigorous yet accessible evaluation.
- Extensive Evaluation and Insights: We benchmark 19 state-of-the-art LLMs on EcomEval, revealing clear strengths and weaknesses across tasks and offering actionable guidance for future model development and deployment in e-commerce.

## 2. Related Work

### 2.1. LLMs for E-Commerce

LLMs show great potential in improving day-to-day online shopping experience (Palen-Michel et al., 2024). For example, LLMs are used in product attribute extraction (Fang et al., 2024; Zhang et al., 2025), user query understanding for better search relevance (Tang et al., 2025; Wang and Na, 2024), and personalised product recommendations (Wang et al., 2024; Xu et al., 2024). To achieve good performance in e-commerce applications, the first e-commerce instruction dataset, EcomInstruct (Li et al., 2023), is proposed with instruction data from various e-commerce tasks. ECInstruct (Peng et al., 2024) covers a wider range of tasks, with each instruction data consisting of an instruction, an input, and an output.

### 2.2. E-commerce Benchmarks

In Table 1, we compare EcomEval with related e-commerce datasets. "Difficulty-aware" indicates whether tasks are stratified by difficulty level,"Multi-cls" denotes whether tasks are categorized into multiple fine-grained classes,and "Real-world" signifies whether these datasets' tasks are mainly from the real world. EcomInstruct-test(Li et al., 2023) primarily draws from open-source datasets. eCeLLM (ECInstruct-test)(Peng et al., 2024) is derived entirely from real-world scenarios, yet it lacks multimodal tasks which are critically important in e-commerce applications. MMECInstruct-test(Ling et al., 2024) addresses this gap. To address the absence of comprehensive evaluation resources for Chinese e-commerce ecosystems, ChineseEcomQA (Chen et al., 2025), a question-answering benchmark is proposed. However, most of the aforementioned evaluation datasets suffer from limited language coverage and overly homogeneous task taxonomies. To benchmark the reasoning and multi-lingual abilities of e-commerce LLMs, Shopping MMLU (Jin et al., 2024) is proposed with tasks covering categories in shopping concept understanding, shopping knowledge reasoning, and user behavior alignment. Nevertheless, Shopping MMLU contains partially synthetic data and lacks multimodal task support. Moreover, none of these datasets incorporate difficulty-tiered task design — a crucial component for fine-grained model evaluation.

Table 1 | EcomEval vs. Peer E-Commerce Datasets

| Dataset | Languages | Tasks | Multimodal | Difficulty-aware | Multi-cls | Real-world |
|---|---|---|---|---|---|---|
| ECInstruct-test | 1 | 10 | ✗ | ✗ | ✓ | ✓ |
| MMECInstruct-test | 1 | 7 | ✓ | ✗ | ✗ | ✓ |
| ChineseEcomQA | 1 | 10 | ✗ | ✗ | ✗ | ✗ |
| EcomInstruct-test | 2 | 12 | ✗ | ✗ | ✗ | ✗ |
| Shopping MMLU | 6 | 57 | ✗ | ✗ | ✓ | ✓ |
| EcomEval(Ours) | 7 | 37 | ✓ | ✓ | ✓ | ✓ |

## 3. EcomEval

In this section, we first introduce the classification logic of the EcomEval task taxonomy. We elaborate on the six primary categories along with their corresponding specific tasks. Then, we describe the definition of the difficulty level. Finally, we provide a detailed explanation of the dataset construction methodology.

### 3.1. Introduction to EcomEval Task Tree

Our e-commerce task tree is hierarchical. Its design philosophy is to cover as many real-world e-commerce scenarios as possible. Specifically, our task tree includes 6 primary categories and dozens of tasks. All tasks under different categories are non-overlapping. The e-commerce tasks in EcomEval mainly come from our internal application scenarios. The rest of the benchmark originates from open-source e-commerce data for which we have provided Southeast Asian multilingual versions through our internal annotation team. The double-layered donut chart in Figure 1 illustrates our e-commerce tasks, with the inner ring representing the six main categories and the outer ring showing the specific tasks.

The six primary categories are: Ecom Question Answering, Shopping Concepts, User Understanding, Shopping Reasoning, Ecom Generation and Ecom Multimodal. Among these, Ecom Question Answering mainly pertains to the user's shopping perspective and includes basic e-commerce knowledge. Shopping Concepts focuses on the merchant's perspective, such as product attributes, features, classifications, etc. User Understanding mainly refers to the comprehension of user behavior in the e-commerce domain. Shopping Reasoning primarily involves tasks such as product recommendation, price calculation, and other reasoning-related activities. Ecom Generation, mainly encompasses content generation related to product information and descriptions. The last category, Ecom Multimodal, tasks under this category combine multimodal and textual information, including product cover image understanding, prohibited product image detection, etc. Additionally, the dataset includes seven languages: Chinese, English, Vietnamese, Thai, Indonesian, Malay, and Portuguese, addressing the lack of low-resource language data in conventional e-commerce datasets. The details of some tasks' example can be found in the Appendix A.

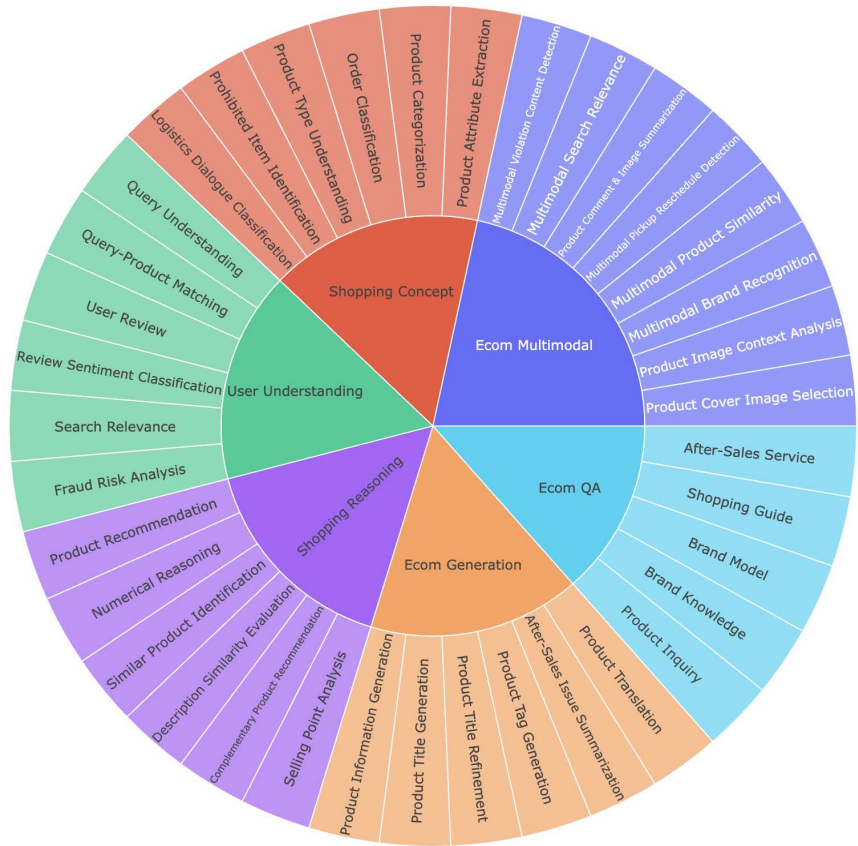

Figure 1 | Overview of the EcomEval benchmark, organized into six primary categories—Ecom Question Answering, Shopping Concepts, User Understanding, Shopping Reasoning, Ecom Generation, and Ecom Multimodal—covering both user- and merchant-oriented tasks. The dataset spans seven languages (Chinese, English, Vietnamese, Thai, Indonesian, Malay, and Portuguese), addressing the scarcity of multilingual resources in e-commerce evaluation.

In addition, to make the task tree more comprehensive, we sample data from some tasks in the eCeLLM and Shopping MMLU datasets, and translate these data from English into low-resource languages. The tasks of product inquiry, product recommendation, similar product identification, review sentiment classification, and product attribute extraction are jointly derived from eCeLLM and Shopping MMLU. Meanwhile, numerical reasoning is sourced exclusively from Shopping MMLU.

3.2. Difficulty Level

Our dataset's tasks come from authentic e-commerce scenarios and encompasse a diverse array of question formats, ranging from multiple-choice questions to open-ended, generation-based tasks. The difficulty levels of the questions vary accordingly. Following the scoring rubric outlined in Appendix B, we evaluate each response from the tested models shown in section 4.2 (ranging from 7B to several hundred billion parameters)on a 0–3 point scale. We then compute average scores of each task's questions and linearly rescale them to a 0–100 point range for easier interpretation and comparison. Based on the average scores of these tasks, we categorize the tasks into three difficulty levels: easy, medium, and hard. Specifically, tasks with an average score below 70 are labeled as hard, those with scores between 70 and 80 are classified as medium, and those exceeding 80 are considered easy. Appendix C provides examples of tasks across different difficulty levels.

3.3. Dataset Construction Methodology

As shown in Figure 2, this paper constructs EcomEval through the following four steps.

Step 1: Online Log Collection. We have collected online logs when users used open-source or closed-source large language models. These logs are categorized into two types, one type is API invoke data, where users solve specific business problems by calling LLM APIs. During a single usage session, users often submit multiple requests related to a particular business scenario, which tend to have similar or even identical prompts. This means that API business data for the same type of task contains different inputs but shares similar instruction contexts. The other type is data from LLM websites. Users request LLM services through a chat interface, and the prompts in this type of data are relatively random.

Step 2: Task Classification. API call queries within the same business usually share the same prompt instruction prefix. Therefore, for such queries, we adopt a prefix clustering method to extract different tasks. Assume that each task requires at least $n$ samples (in this work, $n = 1000$). We take the first $m$ (we try $m = 10, 20, 50$ to extract different tasks) characters from the API call query to obtain prefix_m, then classify the logs based on prefix_m. We assume that questions sharing the same prefix_m belong to the same task and are therefore grouped together accordingly. For clusters containing a large number of queries, we increase $m$ to perform more fine-grained clustering within that cluster.

We obtain several clusters through the prefix clustering method. Then, we employ GPT-4o to automatically generate a task name of approximately 10 words, which is then manually verified for accuracy. This method works well for API data with identical prompts. However, for website data, it is often difficult to classify using prefixes. Therefore, we fine-tune a classification model to categorize website data.

We construct the training dataset for the task classification model based on our internal e-commerce scenarios. To ensure semantic accuracy and annotation quality, the raw corpus is meticulously labeled through manual annotation. The design of the classification model's label integrates clustering analysis results from historical API logs of the platform, along with task type structures from several mainstream open-source e-commerce datasets, ultimately establishing 37 high-frequency and business-representative e-commerce task categories, as shown in Figure 1. Each task category contains 1,000 annotated samples that have undergone cleaning, deduplication, and standardization, ensuring balanced data distribution across all categories. To further enhance the model's adaptability to unknown or emerging user intents, a general "Others" category is introduced to capture novel or marginal requests not covered by the current classification label.

For model training, we select Qwen3-8B-Instruct as the base model, which demonstrates strong semantic understanding and generalization capabilities in e-commerce tasks and multilingual comprehension scenarios. To adapt the model to our classification task, we employ LoRA for parameter

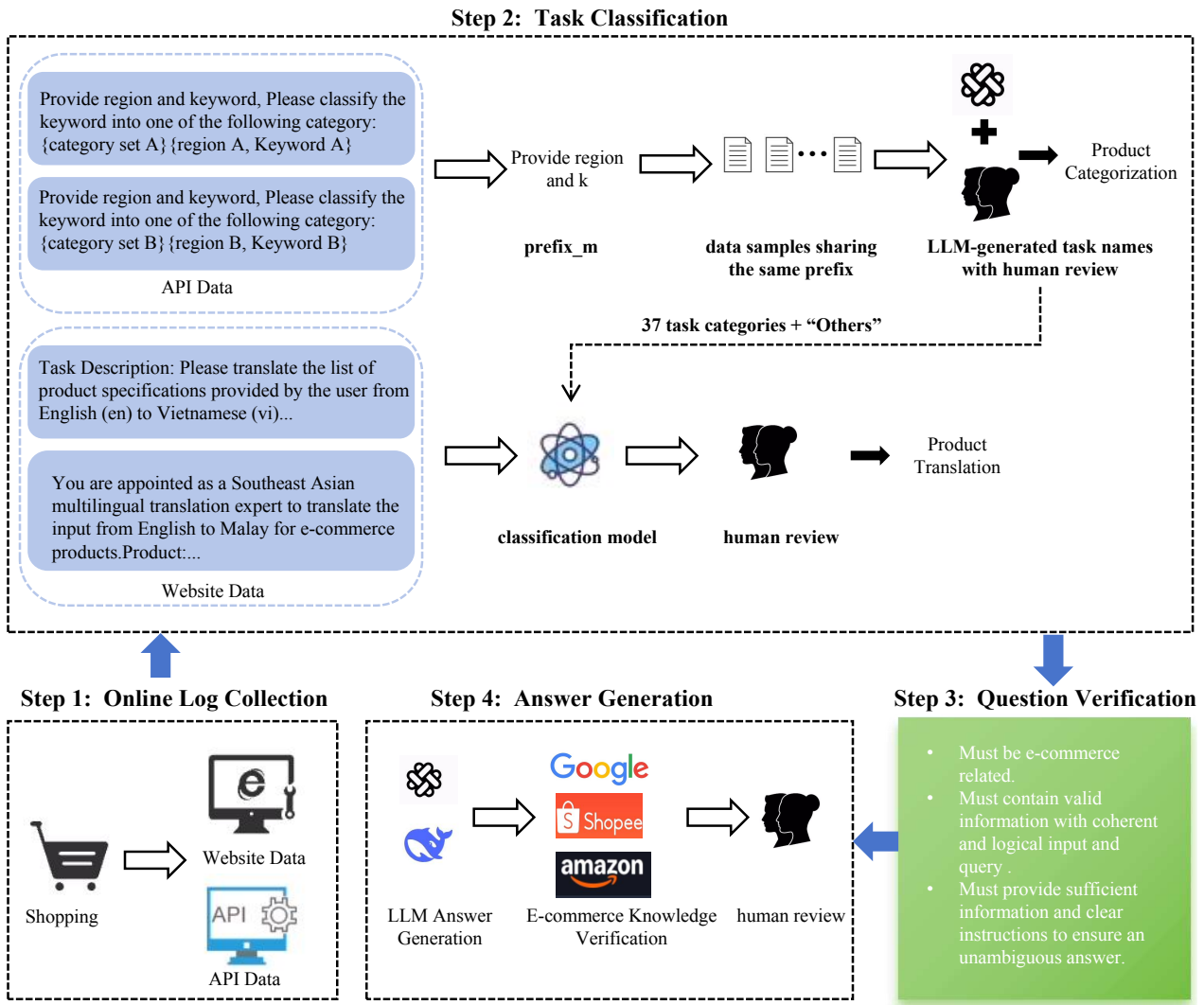

Figure 2 | Overall pipeline of EcomEval dataset construction. The process consists of four stages: (1) collecting API call logs and website queries from LLM usage; (2) clustering API data via prefix grouping and classifying website data with a fine-tuned model to form 37 representative task categories; (3) verifying that sampled questions are e-commerce–relevant, coherent, and unambiguous; and (4) generating and fact-checking answers with LLMs, external sources, and human expert review across multiple languages.

fine-tuning.

We also apply our classification model to categorize the website data, then sample a few data points from each category and manually annotate these samples to further verify the classification accuracy. Finally, we construct our evaluation set by selecting some samples from each category that have been manually confirmed.

Step 3: Question Verification. After collecting data from various categories using the aforementioned methods, we need to further review the quality of the questions. The questions must be e-commerce related and cover scenarios across the pre-sales and after-sales stages. Each question should contain sufficient and meaningful information. Assuming a question consists of input data and an instructional component, the semantics of both components must be coherent, logically consistent, and free of contradictions. Additionally, the information provided and the instructions given in the question must be clear and comprehensive enough for an unambiguous answer. If the sampled data does not meet any of the criteria, it shall be deleted.

Step 4: Answer Generation. For the high-quality questions collected, we use LLMs to generate corresponding answers. For questions requiring specialized knowledge in e-commerce, such as product usage, brand information, or industry-specific content, we further verify the accuracy of the information using popular search engines such as Google, Bing, Baidu, as well as leading e-commerce platforms, including Shopee, Amazon, Lazada, and others. This may involve consulting existing blogs, official e-commerce stores, product pages, and user reviews to ensure factual correctness. Finally, human reviewers examine whether the generated answers follow all the instructions provided in the question and whether the answers are accurate, complete, and free from ambiguity.

In addition, we employ a team of over 50 human experts proficient in the target languages to manually review the quality of questions and answers in step 3 and step 4. These experts come from different countries and know the local culture and language well. They hold diverse educational backgrounds and have worked in the e-commerce field for many years.

## 4. Experiments

### 4.1. Evaluation Method

We have evaluated closed-source and open-source LLMs and Multimodal LLMs (MLLMs) on EcomEval tasks. We adopt LLM-as-a-judge, i.e., GPT-4.1, to evaluate the answers from each LLM/MLLM. LLM judge is asked to compare the model response to the reference answer in relation to the question and assign a score. We attach the 0-3 point scoring rubric in Appendix B. The score is reviewed by expert annotators to ensure the correctness of the evaluation.

### 4.2. Experiment Setup

We evaluate various closed-source and open-source LLMs and MLLMs, including GPT (OpenAI, Hurst et al. (2024)), Gemini (DeepMind), Qwen (Yang et al. (2025), Qwen et al. (2025)), and Llama (Meta, 2025) series. Note that for all open-source models, the instruction finetuned versions are used in our experiments. As mentioned in Section 4.1, each model response is evaluated by the LLM judge on a 0-3 point scoring rubric and then reviewed by expert annotators. For analysis purposes, the 0-3 score is converted linearly to a percentage.

Table 2 | Performance of Closed-Source and Open-Source LLMs on EcomEval Tasks.

| Models | Average | Ecom Generative Ability | User Understanding | Shopping Reasoning | Shopping Concept | Ecom QA |
|---|---|---|---|---|---|---|
| Closed-Source Large Language Models | | | | | | |
| GPT-4o | 69.57 | 63.84 | 66.77 | 67.62 | 77.14 | 72.45 |
| GPT-4o-mini | 66.34 | 58.49 | 64.48 | 66.14 | 75.69 | 66.89 |
| GPT-4.1 | 73.52 | 69.14 | 68.25 | 72.70 | 79.35 | 78.15 |
| GPT-4.1-mini | 70.34 | 65.89 | 65.07 | 70.68 | 78.29 | 71.75 |
| GPT-5 | 73.69 | 68.71 | 68.17 | 74.82 | 79.96 | 76.79 |
| Gemini-2.5-pro | 71.41 | 66.61 | 68.69 | 68.41 | 78.74 | 74.60 |
| Gemini-2.5-flash | 70.27 | 66.06 | 67.68 | 66.74 | 77.52 | 73.36 |
| Average | | 65.54 | 67.02 | 69.59 | 78.10 | 73.43 |
| Open-Source Large Language Models | | | | | | |
| Qwen2.5-14B | 62.77 | 53.60 | 62.79 | 60.09 | 72.09 | 65.28 |
| Qwen2.5-72B | 66.51 | 58.69 | 64.88 | 64.54 | 76.54 | 67.88 |
| Qwen3-14B | 65.80 | 55.06 | 64.19 | 62.70 | 73.39 | 73.64 |
| Qwen3-32B | 67.29 | 58.80 | 65.43 | 65.86 | 74.62 | 71.73 |
| Qwen3-30B-A3B | 63.46 | 53.06 | 61.50 | 63.30 | 69.89 | 69.53 |
| LLaMa3-8B | 56.25 | 29.87 | 61.20 | 55.15 | 72.55 | 62.49 |
| LLaMa3-70B | 63.18 | 52.54 | 61.73 | 62.70 | 73.53 | 65.38 |
| LLaMa4-scout | 62.93 | 54.45 | 63.16 | 59.87 | 72.84 | 64.31 |
| Average | | 52.01 | 63.11 | 61.78 | 73.18 | 67.53 |

4.3. Analysis and Case Study

4.3.1. Overall performance

Overall, proprietary models, i.e., GPT models and Gemini models, perform better than open-source models. Specifically, GPT-5 has the best overall performance with an average score of 73.69%, GPT-4.1 comes close at 73.52%. GPT-5 tops the charts for tasks in the categories of shopping concept and shopping reasoning, whereas GPT-4.1 is the best model for Ecom QA and Ecom Generative Ability. The results show that Qwen3-32B outperforms other open-source LLMs and GPT-4o-mini. Although it is reported by Yang et al. (2025) that Qwen3-32B shows much better performance in general tasks than GPT-4o-mini, it is not the case in EcomEval, with Qwen3-32B (67.29%) performing just slightly better than GPT-4o-mini (66.34%). This shows that the performance in general benchmarks does not generalize to domain-specific tasks, and our benchmark can guide the development of LLMs in the e-commerce domain.

### 4.3.2. Multilingual performance

The average score by language is presented in Table 3. While the gap between the best model and the second-best model is small for tasks in English (0.8%), the difference becomes significant in languages such as Indonesian (3.37%) and Malay (2.28%). Qwen3-32B performs notably better in Chinese, leading Llama4-scout by 6.65%. However, Llama4-scout achieves better results in Malay with 4.06% higher than Qwen3-32B. The variance of the performance shows the importance of having multilingual questions in LLM benchmarks to evaluate the multilingual capabilities of the models. Overall, all models perform better for tasks in English as compared to tasks in low-resource languages.

Table 3 | Multilingual performance of Closed-Source and Open-Source LLMs on EcomEval Tasks.

| Models | English | Indonesian | Malay | Portuguese | Thai | Chinese | Vietnamese |
|---|---|---|---|---|---|---|---|
| Closed-Source Large Language Models | | | | | | | |
| GPT-4o | 72.73 | 62.89 | 64.10 | 67.29 | 67.34 | 71.85 | 65.82 |
| GPT-4o-mini | 69.93 | 59.26 | 63.61 | 66.93 | 62.64 | 66.93 | 64.46 |
| GPT-4.1 | 75.20 | <u>69.82</u> | 67.19 | <u>70.86</u> | 70.48 | 75.74 | <u>70.18</u> |
| GPT-4.1-mini | 72.47 | 67.13 | 65.07 | 67.76 | 67.20 | 71.33 | 66.64 |
| GPT-5 | <u>74.40</u> | 73.19 | 72.23 | 71.81 | 69.48 | <u>74.18</u> | 72.08 |
| Gemini-2.5-pro | 73.40 | 66.96 | <u>69.95</u> | 69.79 | <u>69.91</u> | 72.97 | 69.90 |
| Gemini-2.5-flash | 72.93 | 66.70 | 67.51 | 67.17 | 67.34 | 71.77 | 68.95 |
| Open-Source Large Language Models | | | | | | | |
| Qwen2.5-14B | 66.47 | 55.88 | 59.22 | 62.40 | 59.65 | 66.93 | 59.16 |
| Qwen2.5-72B | 69.47 | 59.34 | 61.82 | 64.43 | 63.93 | 69.69 | 65.82 |
| Qwen3-14B | 69.73 | 57.18 | 61.66 | 63.83 | 64.21 | 70.12 | 62.83 |
| Qwen3-32B | 71.47 | 60.99 | 60.20 | 65.02 | 63.07 | 70.21 | 66.10 |
| Qwen3-30B-A3B | 68.00 | 57.52 | 58.73 | 60.74 | 60.65 | 66.67 | 60.38 |
| LLaMa3-8B | 62.80 | 52.76 | 48.16 | 52.64 | 53.95 | 54.75 | 47.05 |
| LLaMa3-70B | 68.40 | 55.36 | 61.98 | 63.60 | 60.08 | 63.82 | 60.65 |
| LLaMa4-scout | 68.80 | 54.06 | 64.26 | 62.29 | 63.50 | 63.56 | 60.52 |

### 4.3.3. E-commerce Performance

In this section, we discuss the e-commerce performance of the models by analyzing the categories where the models do not perform well. From Table 2, we observe that both closed-source and open-source models perform poorly in e-commerce generative tasks. Particularly, the models underperform in product tag generation and product title generation tasks, missing selling points in the product tag/title generation. Another category where the models do not perform well is user understanding. The models often overlook the information of product categories and targeted buyers when tested with tasks such as query-product matching and search relevance.

Moreover, we observe that in Ecom QA, which includes the tasks of shopping guide and after-sales service (absent in existing e-commerce benchmarks), it is challenging for LLMs to tackle these questions from real-world online shopping scenarios. With LLMs scoring between 62%-78%, there is clear room for improvement in these tasks. These two tasks are representative in e-commerce applications because they reflect the ability of LLMs to guide buyers through pre-sales and post-sales stages. The relatively modest performance in these areas highlights both the incompetence and the importance of strengthening LLM in shopping guide and after-sales tasks. We include the bad cases of the shopping guide and after-sales in Table D3 and Table D4, respectively.

### 4.3.4. Multimodal Performance

Table 4 | Performance of Closed-Source and Open-Source MLLMs on EcomEval Multimodal Tasks.

| Models | Average | PCISel | PICA | MBR | MPS | MPRD | PCISum | MSR | MVCD |
|---|---|---|---|---|---|---|---|---|---|
| Closed-Source Multimodal Large Language Models | | | | | | | | | |
| GPT-4o | 62.44 | 71.33 | 45.89 | 70.33 | 55.01 | 63.11 | 61.67 | 68.16 | 64.03 |
| GPT-4.1 | 76.66 | 78.82 | 81.33 | 73.67 | 67.38 | 80.44 | 86.33 | 70.80 | 74.50 |
| GPT-5 | 77.04 | 81.20 | 83.09 | 77.67 | 62.23 | 82.00 | 87.33 | 71.60 | 71.20 |
| Gemini-2.5-pro | 74.44 | 78.48 | 79.58 | 78.00 | 63.95 | 66.67 | 86.00 | 70.54 | 72.30 |
| Gemini-2.5-flash | 73.27 | 78.82 | 80.28 | 71.67 | 68.07 | 59.78 | 82.67 | 75.04 | 69.82 |
| Average | | 77.73 | 74.03 | 74.27 | 63.33 | 70.40 | 80.80 | 71.23 | 70.37 |
| Open-Source Multimodal Large Language Models | | | | | | | | | |
| Qwen2-VL-7B | 47.29 | 67.93 | 57.82 | 47.67 | 60.16 | 43.33 | 21.67 | 31.92 | 47.78 |
| Qwen2.5-VL-7B | 48.05 | 71.33 | 70.46 | 31.00 | 47.79 | 42.67 | 25.33 | 42.23 | 53.56 |
| Qwen2.5-VL-72B | 60.82 | 71.67 | 79.58 | 57.33 | 63.95 | 56.89 | 37.00 | 53.34 | 66.79 |
| LLaMa3.2-11B-Vision | 37.20 | 43.78 | 59.93 | 28.67 | 31.64 | 25.78 | 25.33 | 38.53 | 43.92 |
| LLaMa4-scout | 59.92 | 75.76 | 70.11 | 81.67 | 47.11 | 43.33 | 51.33 | 56.25 | 53.84 |
| Average | | 66.09 | 67.58 | 49.27 | 50.13 | 42.40 | 32.13 | 44.46 | 53.18 |

From Table 4, we can observe that GPT-5 and GPT-4.1 are the top two models in terms of average scores across all multimodal tasks. GPT-5 has the best performance in 4 tasks, namely product cover image selection (PCISel), product image content analysis (PICA), multimodal pickup reschedule detection (MPRD), and product comment & image summarization (PCISum). Gemini-2.5-flash outperforms other models in the tasks of multimodal product similarity (MPS) and multimodal search relevance (MSR), while GPT-4.1 comes out top for multimodal violation content detection (MVCD). Unexpectedly, LLaMa4-scout attains the highest score for multimodal brand recognition (MBR), surpassing all other models. We highlight that even proprietary models show weak results for the task MPS. When analyzing the results from this category, we find that the models tend to overestimate the similarity of two products by comparing their product titles, overlooking subtle differences in their images. This underscores the importance of enhancing the multimodal capability of MLLM-based shopping assistants to achieve fine-grained product understanding.

## 5. Conclusion

This paper presents EcomEval, a multilingual and multimodal e-commerce benchmark covering 6 primary categories and 37 tasks across 7 languages, providing comprehensive real-world use cases in the e-commerce domain. From the average scores of each question and category, we derive difficulty tags at the question-level and category-level, respectively. The difficulty tags are included in our open-source benchmark. Our benchmark mainly consists of single-turn questions. For future work, multi-turn tasks relevant to online shopping scenarios can be added to assess the multi-turn capability of LLMs in a multilingual setting.

## A. Examples of EcomEval's Tasks

### A.1. Question Answer - Brand Knowledge

#### A.1.1. Question

你是一个品牌合规检测专家，按以下规则处理输入数据
你是一个擅长品牌识别和文本分析的助手。请提取出{title}中产品符合{category_path}特征的自身品牌。

输入内容:[{"category_path": "商品绑定的完整类目路径范围依次缩小","brand":"品牌名" "title": "商品的标题信息"}]

处理规则:

规则一:

步骤1) 在{title}中寻找可能的品牌信息

步骤2) 如果提取到关键词是{brand}的简称、缩写、错误写法、子品牌、母公司、合作伙伴、竞争对手时输出N,没有则输出Y

规则二: 只使用{category_path}从{title}中提取品牌

步骤1) 逐步分析{category_path}中每个类目的特征

步骤2) 在{title}中寻找主营产品符合该类目特征的品牌

步骤3) 排除验证：当候选词同时满足以下条件时保留

- 非产品型号特征
- 非营销术语（排除旗舰版/专业版等）
- 候选品牌与当前产品不存在配件/耗材类关系

步骤4) 这里禁止判断与{brand}的关系，单独判断候选词是否是明确的品牌，如果是独立品牌则输出N。未找到明确品牌信息则输出Y

注意：

- 必须严格按照规则中每个步骤进行处理,
- 分别处理规则一和规则二并保存处理结果
- 只有规则一和规则二的判断结果都是Y时，res输出Y，其余情况输出N
- 若提取到到品牌信息不确定是否是独立品牌则在brand中额外输出+可能是品牌+

输出要求：

- 用JSON数组格式严格输出[{"res":"Y/N","brand":"仅N时存在,标题中提取到的品牌信息","explain":"每一步完整的推理过程"}]
- explain需用中文详细说明分析步骤
- 禁止输出除json外额外内容

现在处理这个输入：

{"title": "Used shuttlecock Gred A, B & C", "category_path": ["Sports & Outdoors", "Sports & Outdoor Recreation Equipments", "Badminton", "Shuttlecocks"], "brand": "Felet"}

### A.1.2. Reference Answer

[{"res": "Y", "explain": "规则一分析：标题'Used shuttlecock Gred A, B & C'中未提取到品牌信息Felet，规则一输出Y。规则二分析：各类目在标题中没有提取到任何品牌信息，规则二输出Y。最终两规则均Y故res=Y"}]

## A.2. Shopping Concepts - Prohibited Item Identification

### A.2.1. Question

You are a safety policy enforcement model. Your task is to determine whether the query violates safety policies (e.g., contains keywords of Adult content/product, Sensitive content/product, Prohibited items such as guns, drug, vape, and illegal brand, or Underwear and Intimate Apparel including all types of bras, panties, shapewear, bikini, lingerie, regardless of style or presentation).

Query: panty for kids girls 1 to 2 years old

If the query violates safety policies, output 1. Otherwise, output 0. Do not explain.

Output:

### A.2.2. Reference Answer

### A.3. User Understanding - Review Sentiment Classification

#### A.3.1. Question

Please classify the delivery experience and sentiment in the customer's comment below.
- Delivery Experience: Describe rider behavior, delivery speed, overall delivery quality, etc.
- Sentiment: Choose only one: Positive, Neutral, or Negative.
Customer comment: "User Tips:Buy it now, ypu won't regret it the product is good
Packaging:The product is safe coz' the item is packaged is well.
Beauty Profile:Perfect for my skin tone
Received item safe coz'it's packed well. Thank you seller and also to rider."
Respond in this format:
Delivery Experience: <short description>
Sentiment: <Positive/Neutral/Negative>

#### A.3.2. Reference Answer

Delivery Experience: The delivery experience was positive, with the rider being acknowledged and the item being received safely due to good packaging.
Sentiment: Positive

### A.4. Shopping Reasoning - Product Recommendation

#### A.4.1. Question

I provide a product description '1. Butt Lifter Slimming High Waist Girdle Corset Long Shaper Girdle Pants Plus Size Girdle Shapewear BengkungColour Black Nude Size M L XL 2XL 3XL 4XL Measurement Size M L Waist 65 80 CM 26 32 inch Weight 50 65 KG Size XL XXL Waist 80 100 CM 32 40 inch Weight 65 80 KG Size 3XL Waist 100 120 CM 40 47 inch Weight 80 100 KG Size 4XL Waist 120 135 CM 47 53 inch Weight 100 115 KG Material Spandex and Nylon Slim your Tummy Waist and Bottom and wear figure hugging clothes with new confidence Breathable and Comfortable material Wear All day along 2. Pants Girdle Plus Size Corset Girdle Slimming Girdle Shapewear Borong Bengkung High Waist Girdle t Welcome to my shop nMalaysia local seller nFollow the store to get more real time data of store products nWe ship daily order today and ship immediately n100 brand new high quality product n nFeatures nColor Black Beige nSize M L XL XXL 3XL 4XL n1 on waist spiral steel bone to prevent curling n2 breathable comfortable stretchable healthy no smell perfect design n3 butt lifter with tummy control waist trainer waist slimming n4 high waist butt lift panties n5 corrective slim underwear n6 Suitable season four seasons nSize of the prodcut Same as the picture shown nSize Waist Weight nM L 20 28 40 50KG nXL 2XL 26 31 50 60KG n3XL 30 35 60 70KG n4XL 33 40 70 80KG nPackage Includes 1pcs Women Control Panties i cn 11134208 7ras8 m4r51le1xjrd69

1 000 i cn 11134208 7ras8 m4r51le1yybt0d 1 000 i cn 11134208 7ras8 m4r51le20cw9b1 1 000 i cn 11134 3. ANGELMOMMY Double-Breasted High-Waist Abdomen Tucking Underwear Body-Shaping Belly-Shaping Pants Women Hip Lifting 039 t Welcome to angelmommy os where we strive to provide our customers an array of products that satisfy demand for quality and efficiency Buy our products in a worry free manner as we guarantee 100 authenticity Shopping can t get any easier than this so start today n n Product features n Six advantages n Focus on abdominal waist n Improve abdominal relaxation n Correction of pelvic expansion n Shape sexy curves n Three dimensional hip lift n Shape legs n n Product effect n Flat Belly flat belly n Waist sculpting tighten fat n Butt lift curve butt lift n Slim legs shape legs n n Product details n Three breasted buckle for more shape n Upgraded three breasted belly pants n Through the continuous pressurization of the buckle n Get rid of belly fat n Make your belly thinner and thinner n Bring you back to your graceful posture n n Product si' for you. Please analyze the product names mentioned and recommend up to 10 complementary items to enhance the user's purchase experience. Focus on relevant and useful complementary items that naturally pair with the main product. Ensure the recommendations follow these rules:

1. Exclude included accessories.
2. Ensure compatibility (brand, model, category).
3. Avoid items of the same type as 'Shapewear'.
4. Exclude shipping/packaging items.
5. Rank by purchase co-occurrence.
6. Avoid sets containing 'gurdle'.

Generate keyword list for the recommended items, using 'Malaysia Malay' language whenever possible: 'keywords': Various related keywords for practical items.
Ensure diversity and relevance of the keyword while avoiding duplication. Output in JSON format: "keywords": [...], do not return any other texts

A.4.2. Reference Answer

"keywords": ["baju dalam wanita", "bra tanpa dawai", "bralette selesa", "pakaian dalam seamless", "stokin wanita", "pakaian tidur wanita", "losyen pelangsing badan", "pencuci pakaian dalam", "pakaian senaman wanita", "minyak urut pelangsing"]

A.5. Generation Ability - Product Title Translation

A.5.1. Question

Task Description: Please translate the list of product specifications provided by the user from English (en) to Vietnamese (vi). Input Format: 1. Multiple specifications enclosed in square brackets (e.g., [Specification 1, Specification 2, Specification 3]), with each specification separated by a semicolon (;).

Translation Requirements: 1. Format Preservation: The output format should be similar to the input format, maintaining the original list structure, with semicolons used as separators. 2. Output Content: Only output the translated specifications, without explanations or any additional content. 3. Accuracy: Ensure the translations are as accurate as possible.

[Anti-pilling: Green/White+Pink/White+Blue/White; Anti-pilling: Green/White+Pink/White+White/Yellow; Anti-pilling: White/Card+White/Blue+White/Yellow; Anti-pilling: White/Blue+White/Yellow+Pink/White; Anti-pilling: Green/White+White/Blue+White/Card; Anti-pilling: Green/White+Blue/White+White/Yellow+Wh Anti-pilling: Blue/White+White/Yellow+White/Card+Pink/White; Anti-pilling: Green/White+Blue/White+Whit Anti-pilling: Green/White+Green/White+White/Blue+White/Blue; Anti-pilling: Green/White+Green/White+Wh

Anti-pilling: White/Yellow+White/Yellow+Pink/White+Pink/White; 6 styles, 1 pair each; 3 pairs, random if no note; 4 pairs, random if no note]

A.5.2. Reference Answer

[Chống xù: Xanh/Trắng+Hồng/Trắng+Xanh/Trắng; Chống xù: Xanh/Trắng+Hồng/Trắng+Trắng/Vàng; Chống xù: Trắng/Thẻ+Trắng/Xanh+Trắng/Vàng; Chống xù: Trắng/Xanh+Trắng/Vàng+Hồng/Trắng; Chống xù: Xanh/Trắng+Trắng/Xanh+Trắng/Thẻ; Chống xù: Xanh/Trắng+Xanh/Trắng+Trắng/Vàng+Trắng/Thẻ Chống xù: Xanh/Trắng+Trắng/Vàng+Trắng/Thẻ+Hồng/Trắng; Chống xù: Xanh/Trắng+Xanh/Trắng+Trắng/Xan Chống xù: Xanh/Trắng+Xanh/Trắng+Trắng/Xanh+Xanh/Trắng; Chống xù: Xanh/Trắng+Xanh/Trắng+Trắng/Và Chống xù: Trắng/Vàng+Vàng/Trắng+Hồng/Trắng+Hồng/Trắng; 6 mẫu, mỗi mẫu 1 đôi; 3 đôi, ngẫu nhiên nếu không ghi chú; 4 đôi, ngẫu nhiên nếu không ghi chú]

A.6. Ecom Multimodal

A.6.1. Question

Analyze the image and identify the brand of the product. If the brand matches one from the list below, return only the exact brand name(s). If no match is found or the brand is not in the list, return "-".

Accepted brands: Calvin Klein, Chanel, John Deere, Levi's, Nike, Tommy Hilfiger, Zimmermann, Lacoste, Oakley, Osklen, Burberry, Estee Lauder, Giorgio Armani, Gucci, Hermés Paris, Hugo Boss, Lancôme, Louis Vuitton, New Balance, Prada, Puma, Ralph Lauren, Ralph Lauren Fragrances, Tiffany & Co., Valentino, Versace, Yves Saint Laurent, Abercrombie, Adidas, Casio, Coach, Converse, Denizen, Diesel, Dockers, JanSport, Jimmy Choo, Jordan, Kiehl's Since 1851, Kipling, Maison Margiela Fragrances, Marc Jacobs, Maybelline New York, Michael Kors, New Era, Nioxin, Supreme, The North Face, Timberland, Under Armour, Vans, Victoria's Secret, Viktor&Rolf, Ray-Ban, Vogue Eyewear, Persol, Oliver Peoples, Arnette, Costa Del Mar, Emporio Armani, Dolce & Gabbana, Miu Miu, Ferragamo, Bvlgari, DKNY, Fendi, Mizuno, Asics, CeraVe, Dior, Azzaro, Bozzano, David Beckham, Dickies, Gabriela Sabatini, Harley-Davidson, Joop!, Kate Spade, Mugler, Redken, SkinCeuticals, Urban Decay, WaterPik, G-Shock, Vichy, Flamengo, Bayer, São Paulo, Palmeiras, Corinthians, Santos, Atlético-MG, Athletico-PR, Fluminense, Botafogo, Fortaleza, Grêmio, Bahia, Internacional, América-MG, Vasco, Manchester City, Real Madrid, Bayern Munich, Liverpool, Paris Saint-Germain, Inter Milan, Chelsea, Barcelona, Manchester United, Juventus, Rolex, Olympikus, Crocs, Zara, Vick, Wahl, Kenner, Loewe, Celine, Montblanc, Dyson, Balenciaga, Panerai, Van Cleef & Arpels, Bottega Veneta, Chloe, Piaget, Cartier, IWC, Pandora, Carolina Herrera, Barbie, Stanley, Bobbie Goods, Fila, Minecraft, Havaianas, PopMart/Labubu. Answer can only be the brand or "-", nothing else.

Table B1 | Scoring Rubric for Model Responses

| Score | Description |
|---|---|
| 3 | The response is entirely correct. |
| 2 | The response is nearly correct but has flaws. |
| 1 | The response is incorrect. |
| 0 | The response is off-topic or violates safety guidelines. |

### A.6.2. Input Image

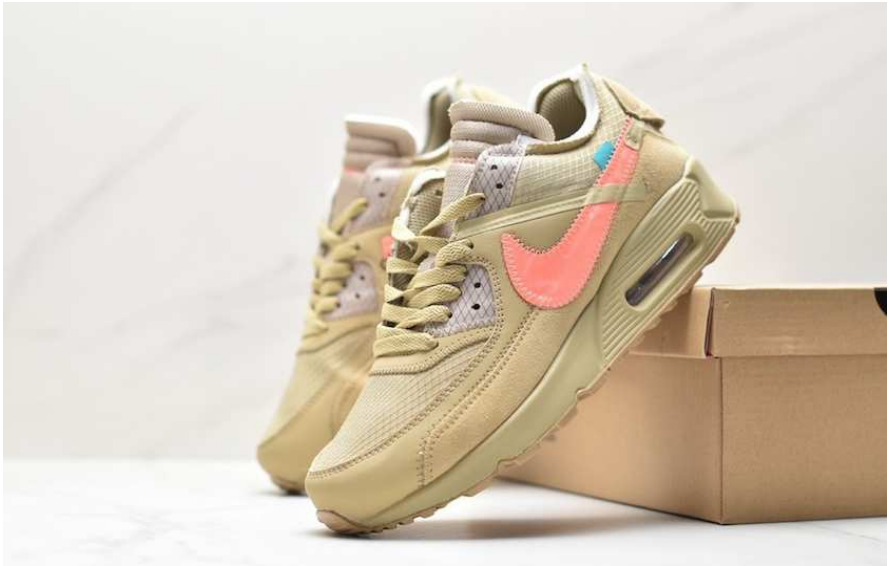

### A.6.3. Reference Answer

Nike

## B. LLM Judge Scoring Rubric

The scoring rubric used in our experiments is presented in Table B1.

## C. Tasks across different difficulty levels

The examples of tasks across different difficulty levels are shown in Table C1. We observe that easy tasks typically assess the capabilities of a model in a straightforward manner, where answers can be derived through simple reasoning from the given information. Most medium-difficulty tasks evaluate the model's grasp of e-commerce-specific knowledge, including product usage and features tailored to Southeast Asian markets, as well as regionally prevalent brand models. Beyond this foundational knowledge, we also assess the model's understanding of after-sales service — such as returns, exchanges, and logistics — testing not only its knowledge of the products themselves, but also its ability to reason through real-world post-purchase processes. Hard tasks not only evaluate the model's mastery of

Table C1 | Examples of tasks across different difficulty levels

| task | key challenges | average | difficulty level |
| --- | --- | --- | --- |
| order classification | Given an order name, output the category to which the order belongs. | 86.89 | easy |
| review sentiment classification | Given the user review, infer the user's sentiment type: negative, neutral, or positive. | 81.16 | easy |
| product categorization | Given the product name and the list of all product categories, output the correct category for the product. | 73.46 | medium |
| similar product identification | Determine the relationship between Product 1 and Product 2 and whether they are similar. | 73.73 | medium |
| product title refinement | Require the model to process input based on complex instructions, and generate a revised title in the specified format and language. | 54.57 | hard |
| search relevance assessment | This task requires assessing the search relevance scores of multiple products within a single query. The output format follows strict specifications, and the data classification logic across different relevance levels is complex, with detailed and precise rules. | 63.6 | hard |

Southeast Asian e-commerce knowledge, but also demand its ability to follow complex instructions to accurately process inputs and generate appropriate outputs.

Additionally, we assign a difficulty level to each individual question based on its average performance across all models. The detailed difficulty labels for each question can be found in the "difficulty level" column of our open-sourced dataset.

## D. Bad Case Analysis

### D.1. User Understanding

We included the bad case in Table D1.

Table D1 | Bad Case in User Understanding Category

| |
| --- |
| Category: User Understanding |
| Question: A user would like to find a product with the keyword 'brass tray rectangular'. What is the relation between the user's intended products and the product boweiwj Stainless Steel Towel Tray Storage Tray Tray Dish Plate Tea Tray Fruit Trays Cosmetics Jewelry Organizer Gold Oval Tray 9in"?<br>A. substitute<br>B. irrelevant<br>C. exact match<br>D. complement |
| Correct answer: A |

Gemini-2.5 Pro Response:
Let's break this down:

User's intended product:
* **Material:** Brass
* **Type:** Tray
* **Shape:** Rectangular

Product 'boweiwj Stainless Steel Towel Tray... Gold Oval Tray 9in':
* **Material:** Stainless Steel
* **Type:** Tray (multiple uses listed)
* **Shape:** Oval
* **Color:** Gold

Comparison:
1. **Material:** User wants "brass," product is "stainless steel." (Mismatch)
2. **Type:** User wants "tray," product is a "tray." (Match)
3. **Shape:** User wants "rectangular," product is "oval." (Mismatch)
4. **Color/Appearance:** Product is "gold" colored, which can sometimes be visually similar to brass, but "brass" is a specific material, not just a color.

Considering the options:
* **A. substitute:** A substitute should be able to perform the same function and be reasonably similar. While it's a tray, the material and shape are different. The gold color might make it a *visual* substitute for some, but "brass" specifies a material. It's a weak substitute at best due to material and shape differences.
* **B. irrelevant:** The product is a tray, which is what the user is looking for in general. However, it fails on two key specific attributes: material (brass vs. stainless steel) and shape (rectangular vs. oval). If the user *specifically* needs brass and rectangular, then this stainless steel oval tray is not relevant to those specific needs.
* **C. exact match:** Clearly not an exact match due to differences in material and shape.
* **D. complement:** This product is not something typically used *with* a brass rectangular tray; it's an alternative type of tray.

The product is a tray, so it's not *completely* unrelated. However, the user has specified "brass" (a material) and "rectangular" (a shape). The offered product is "stainless steel" (a different material) and "oval" (a different shape). Because two of the three explicit keywords describing the desired item are mismatched, the product is not a good fit.

While it is a tray, the deviation in *specified* material and shape makes it quite different from the user's request. If a user explicitly asks for "brass" and "rectangular," offering "stainless steel" and "oval" is more likely to be perceived as irrelevant to their specific need than a suitable substitute.

Therefore, the most appropriate answer is **B. irrelevant** because it fails to meet the key specific attributes requested by the user (material and shape), even though it matches the general product category (tray).

| |
|---|
| The final answer is $\boxed{B}$ |
| LLM Judge: <br> The model response chooses "irrelevant" (B) rather than the correct answer "substitute" (A), despite a detailed comparison; thus, it is incorrect. |
| Final Score: 1 |

D.2. Ecom Generative Ability

We included the bad case in Table D2.

Table D2 | Bad Case in Ecom Generative Ability Category

| |
|---|
| Category: Ecom Generative Ability |
| Question: **Task:** <br> Integrate 'value' into Eucerin Skin Calming Cream / Skin Calming Lotion / Itch Relief Intensive Calming Lotion **without modifying origin content**. Output optimized title only. <br><br> ─── <br><br> **Processing Pipeline** <br><br> **1. Value Preprocessing** <br> - **Preservation Rules**: <br> - Retain ALL original symbols/emojis/formatting ( <...>, 3.5g, 20241014) <br> - Protect multilingual terms in their original form <br><br> - **Term Construction**: <br> - **SPF-exclusive combination**: <br> • **Only** combine 'key_info'+'value' **if** 'key_info="SPF"' (SPF+30→SPF30) <br> • **All other key_info**: Use raw 'value' directly without prefix <br> - Merge variants/ranges: <br> • Overlaps: '30,spf30+' → 'SPF30+' <br> • Ranges: '50ml+100ml' → '50/100ml' <br> • Deduplicate **per category** (SPF50+SPF50→SPF50) <br><br> - **Language Normalization**: <br> - Auto-translate to match adjacent terms' dominant language <br> - Preserve technical codes (PA++++/SPF50) across languages <br><br> **2. Insertion Validation** <br> - **Mandatory Skip Conditions**: <br> - Existing equivalent **in any language** (保湿=Moisturizing) |

- Logical contradictions (Non-Oily vs Oily)
- Full value/key_info already present (skip if "SPF30" exists when inserting SPF30)

**3. Context-Aware Insertion**
- **Priority Positions**:
1. Adjacent to primary product type (e.g., _Sunscreen_ SPF50)
2. Between high-value term & benefit statement
3. Title end (last resort)

- **Linguistic Alignment**:
- Match grammar structure & language flow:
• Insert "保湿" not "Moisturizing" in Chinese-dominant segments
• Mirror adjective/noun order of surrounding text

- **Technical Formatting**:
- Cluster related specs (SPF50/PA++++ Protection)
- Maintain original casing except technical codes

**4. Absolute Constraints**
- NEVER:
• Remove/modify origin content
• Add conjunctions (with/for/by)
• Repeat existing specs in any language
• Insert 'key_info' text (except SPF technical codes)
• Break protected patterns (<Cosmetics>, 3.5g)

---

**Examples**
Valid:
• 'key_info:SPF + 30' → "Sunscreen SPF30 Cream"
• "Sunscreen SPF50/PA++++ 100ml" (clustered specs)
• 'key_info:SkinType + All skIN' → "Lotion All Skin Face Cream" (no "SkinType" prefix)

Invalid:
• "Moisturizing 保湿Cream" (cross-language duplicate)
• "For SPF50 Milk" (illegal preposition)

---

**Output Format**
Return ONLY the optimized title without explanations.

---

**Input**:
- title: Eucerin Skin Calming Cream / Skin Calming Lotion / Itch Relief Intensive Calming Lotion
- key_info: Body Care Benefits
- value: Tender care

| |
|---|
| Reference answer: Eucerin Skin Calming Cream / Skin Calming Lotion / Itch Relief Intensive Calming Lotion Tender Care |
| GPT-4.1 Response: Eucerin Skin Calming Cream Tender care / Skin Calming Lotion Tender care / Itch Relief Intensive Calming Lotion Tender care |
| LLM Judge: The model response incorrectly splits and inserts "Tender care" after each variant, altering the original grouped structure rather than adding the value at the title end, thus violating the constraint to not modify origin content. |
| Final Score: 1 |

### D.3. Ecom Question Answering

#### D.3.1. Shopping Guide

We included the bad case in Table D3

Table D3 | Bad Case in Task Shopping Guide

| |
|---|
| Category: Ecom QA - Shopping Guide |
| Question: Which feature should be considered for comfort and fit in a formal suit?<br>A. Color and patterns<br>B. Trousers cut and break<br>C. Removing or adding padding to shoulders<br>D. Number of buttons and lapel type |
| Correct answer: C |
| GPT-5 Response: B. Trousers cut and break<br>These directly affect how the suit sits and moves on your body, impacting comfort and overall fit. |
| LLM Judge: The model response chooses B, which relates to fit but not comfort in the same direct way as shoulder padding (C), which is more critical; thus, it's nearly correct but not the best answer. |
| Final Score: 1 |

#### D.3.2. After-sales Service

We included the bad case in Table D4

Table D4 | Bad Case in Task After-Sales Service

| |
|---|
| Category: Ecom QA - After-Sales Service |
| Question: Which practice involves refurbishing, remanufacturing, and responsible disposal? <br> A. Extended Producer Responsibility (EPR) <br> B. Adopting Sustainable Practices <br> C. Collaborating with Specialized Logistics Providers <br> D. Working with Third-Party Recyclers |
| Correct answer: B |
| GPT-5 Response: C. Collaborating with Specialized Logistics Providers |
| LLM Judge: The model response incorrectly identifies option C, while the correct answer is B; thus, the response is incorrect. |
| Final Score: 1 |